# Topological Value Iteration Algorithms

**Peng Dai**                                                                DAIPENG@CS.WASHINGTON.EDU
*Google Inc.*
*1600 Amphitheatre Pkwy*
*Mountain View, CA 94043*
*USA*

**Mausam**                                                                  MAUSAM@CS.WASHINGTON.EDU
**Daniel S. Weld**                                                            WELD@CS.WASHINGTON.EDU
*Department of Computer Science and Engineering*
*University of Washington*
*Seattle, WA 98195*
*USA*

**Judy Goldsmith**                                                          GOLDSMIT@CS.UKY.EDU
*Department of Computer Science*
*University of Kentucky*
*Lexington, KY 40508*
*USA*

## Abstract

Value iteration is a powerful yet inefficient algorithm for Markov decision processes (MDPs) because it puts the majority of its effort into backing up the entire state space, which turns out to be unnecessary in many cases. In order to overcome this problem, many approaches have been proposed. Among them, ILAO* and variants of RTDP are state-of-the-art ones. These methods use reachability analysis and heuristic search to avoid some unnecessary backups. However, none of these approaches build the graphical structure of the state transitions in a pre-processing step or use the structural information to systematically decompose a problem, whereby generating an intelligent backup sequence of the state space. In this paper, we present two optimal MDP algorithms. The first algorithm, *topological value iteration* (TVI), detects the structure of MDPs and backs up states based on topological sequences. It (1) divides an MDP into strongly-connected components (SCCs), and (2) solves these components sequentially. TVI outperforms VI and other state-of-the-art algorithms vastly when an MDP has multiple, close-to-equal-sized SCCs. The second algorithm, *focused topological value iteration* (FTVI), is an extension of TVI. FTVI restricts its attention to connected components that are *relevant for solving the MDP*. Specifically, it uses a small amount of heuristic search to eliminate provably sub-optimal actions; this pruning allows FTVI to find smaller connected components, thus running faster. We demonstrate that FTVI outperforms TVI by an order of magnitude, averaged across several domains. Surprisingly, FTVI also significantly outperforms popular 'heuristically-informed' MDP algorithms such as ILAO*, LRTDP, BRTDP and Bayesian-RTDP in many domains, sometimes by as much as two orders of magnitude. Finally, we characterize the type of domains where FTVI excels — suggesting a way to an informed choice of solver.

## 1. Introduction

Markov Decision Processes (MDPs) (Bellman, 1957) are a powerful and widely-adopted formulation for modeling autonomous decision making under uncertainty. For instance, NASA researchers use MDPs to model the next-generation Mars rover planning problems (Bresina, Dearden, Meuleau, Ramkrishnan, Smith, & Washington, 2002; Feng & Zilberstein, 2004; Mausam, Benazera, Brafman, Meuleau, & Hansen, 2005; Meuleau, Benazera, Brafman, Hansen, & Mausam, 2009). MDPs are also used to formulate the military operations planning (Aberdeen, Thiébaux, & Zhang, 2004) and coordinated multi-agent planning (Musliner, Carciofini, Goldman, E. H. Durfee, & Boddy, 2007), etc.

Classical dynamic programming algorithms, such as value iteration (VI), solve an MDP optimally by iteratively updating the value of every state in a fixed order, one state per iteration. This can be very inefficient, since it overlooks the graphical structure of a problem, which can provide vast information about state dependencies.

During the past decade researchers have developed heuristic search algorithms that use reachability information and heuristic functions to avoid some unnecessary backups. These approaches, such as improved LAO* (ILAO*) (Hansen & Zilberstein, 2001), LRTDP (Bonet & Geffner, 2003b), HDP (Bonet & Geffner, 2003a), BRTDP (McMahan, Likhachev, & Gordon, 2005) and Bayesian RTDP (Sanner, Goetschalckx, Driessens, & Shani, 2009), frequently outperform value iteration. On some problems, however, heuristic search algorithms offer little benefit and it is difficult to predict when they will excel. This raises an important, open question, "What attributes of problems and problem domains make them best suited for heuristic search algorithms?"

In this paper we present two algorithms that solve MDPs optimally and speed up the convergence of value iteration: *topological value iteration* (TVI) (Dai & Goldsmith, 2007) and *focused topological value iteration* (FTVI) (Dai, Mausam, & Weld, 2009b). TVI makes use of the graphical structure of an MDP. It performs Bellman backups in a more intelligent order after performing an additional topological analysis of the MDP state space. TVI first divides an MDP into strongly connected components (SCCs) and then solves each component sequentially in topological order. Experimental results demonstrate significant performance gains over VI and, surprisingly, over heuristic search algorithms (despite TVI not using reachability information itself) in a specific kind of domain – one that has multiple, close-to-equal-sized SCCs.

TVI is very general, as it is *independent* of any assumptions on the start state and can find the optimal value function for the entire state space. However, many benchmark problems cannot be broken into roughly equal-sized SCCs, leaving TVI's performance no better (or often worse, due to the overhead of generating SCCs) than other MDP algorithms. For instance, many domains (e.g., Blocksworld) have reversible actions. Problems from these domains that have most of the states connected by reversible actions end up being in one (large) SCC, thus, eliminating the benefit of TVI.

FTVI addresses the weaknesses of TVI. It first performs a phase of heuristic search and eliminates provably sub-optimal actions found during the search. Then it builds a more informative graphical structure based on the remaining actions. We find that a very short phase of heuristic search is often able to eliminate many actions leading to an MDP structure that is amenable to efficient, topology-based solutions.

We evaluate FTVI across several benchmark domains and find that FTVI outperforms TVI by significant margins. Surprisingly, we also find that FTVI outperforms other state-of-the-art heuristic search algorithms in most of the domains. This is unexpected, since common wisdom dictates that heuristic-guided search is much faster than all-state dynamic programming. To better understand this big improvement, we study the convergence speed of algorithms on a few problem features. We discover two important features of problems that are hard for heuristic search algorithms: smaller number of goal states and long search distance to the goal. These features are commonly found in many domains, e.g., Mountain car (Wingate & Seppi, 2005) and Drive (Bonet, 2006). We show that, in such domains, FTVI outperforms heuristic search in convergence speed by an order of magnitude on average, and sometimes by even two orders of magnitude.

Comparing with the previous conference versions (Dai & Goldsmith, 2007; Dai et al., 2009b), this paper makes several significant improvements: (1) We add a convergence test module in the search phase of FTVI. With the module, FTVI works as good as the best heuristic search algorithms in domains where it used to be significantly outperformed. (2) We perform extensive empirical study on both TVI (Figures 2 and 3 are new) and FTVI (Figure 5 is new, and we added the Blocksworld domain). (3) We describe TVI and FTVI in a consistent way and improve the pesudo-codes. (4) We add the convergence proof of TVI (Theorem 2).

The outline of the rest of the paper is as follows: Section 2 formally defines MDPs, and reviews algorithms that solve MDPs. Section 3 describes the topological value iteration algorithm, and compares it empirically with other algorithms on a special MDP domain. Section 4 introduces the focused topological value iteration algorithm and provides a thorough empirical evaluation. We present related work in Section 5 and conclude in Section 6.

## 2. Background

We provide an overview of Markov decision process (MDP) and dynamic programming algorithms that solve an MDP.

### 2.1 Markov Decision Processes for Planning

AI researchers typically use MDPs to formulate fully-observable probabilistic planning problems. An MDP is defined as a five-tuple $\langle \mathcal{S}, \mathcal{A}, Ap, T, C \rangle$, where

- $\mathcal{S}$ is a finite set of discrete states.

- $\mathcal{A}$ is a finite set of all applicable actions.

- $Ap : \mathcal{S} \rightarrow \mathcal{P}(\mathcal{A})$ is the applicability function. $Ap(s)$ denotes the set of actions that can be applied in state $s$. $\mathcal{P}(\mathcal{A})$ is the power set of the set of actions.

- $T : \mathcal{S} \times \mathcal{A} \times \mathcal{S} \rightarrow [0, 1]$ is the transition function describing the effect of an action execution.

- $C : \mathcal{S} \times \mathcal{A} \rightarrow \mathbb{R}^+$ is the cost of executing an action in a state.

The agent executes its actions in discrete time steps. At each step, the system is at one distinct state $s \in \mathcal{S}$. The agent can execute any action $a$ from a set of *applicable actions* $Ap(s) \subseteq \mathcal{A}$, incurring a cost of $C(s, a)$. The action takes the system to a new state $s'$ stochastically, with probability $T_a(s'|s)$.

The *horizon* of an MDP is the number of steps for which costs are accumulated. We concentrate on a special set of MDPs called *stochastic shortest path* (SSP) problems. Despite its simplicity, SSP is a general MDP representation. Any infinite-horizon, discounted-reward MDP can be easily converted to an SSP problem (Bertsekas & Tsitsiklis, 1996). The horizon in such an MDP is *indefinite*, i.e., finite but unbounded, and the costs are accumulated with no discounting. There are two more components of an SSP:

- $s_0$ is the initial state.

- $\mathcal{G} \subseteq \mathcal{S}$ is the set of sink goal states. Reaching any one of $g \in \mathcal{G}$ terminates an execution.

The cost of an execution is the sum of all costs along the path from $s_0$ to the first goal state encountered.

We assume *full observability*, i.e., after executing an action and transitioning stochastically to a next state as governed by $T$, the agent has full knowledge of the state. A policy, $\pi : \mathcal{S} \rightarrow \mathcal{A}$, of an MDP is a mapping from the state space to the action space, indicating which action to execute at each state. To solve the MDP we need to find an *optimal policy* ($\pi^* : \mathcal{S} \rightarrow \mathcal{A}$), a probabilistic execution plan that reaches a goal state with the minimum expected cost. We evaluate any policy $\pi$ by its *value function*, the set of values that satisfy the following equation:

$$V^\pi(s) = C(s, \pi(s)) + \sum_{s' \in \mathcal{S}} T_{\pi(s)}(s'|s) V^\pi(s').$$ (1)

Any optimal policy must satisfy the following system of *Bellman equations*:

$$V^*(s) = 0 \text{ if } s \in \mathcal{G}, \text{ else}$$
$$V^*(s) = \min_{a \in Ap(s)} \left[ C(s, a) + \sum_{s' \in \mathcal{S}} T_a(s'|s) V^*(s') \right].$$ (2)

The corresponding optimal policy can be extracted from the value function:

$$\pi^*(s) = argmin_{a \in Ap(s)} \left[ C(s, a) + \sum_{s' \in \mathcal{S}} T_a(s'|s) V^*(s') \right], \forall s \in \mathcal{S} - \mathcal{G}.$$ (3)

Given an implicit optimal policy $\pi^*$ in the form of its optimal value function $V^*(\cdot)$, the *Q-value* of a state-action pair $(s, a)$ is defined as the value of state $s$, if an immediate action $a$ is performed, followed by $\pi^*$ afterwards. More concretely,

$$Q^*(s, a) = C(s, a) + \sum_{s' \in \mathcal{S}} T_a(s'|s) V^*(s').$$ (4)

Therefore, the optimal value function can be expressed by:

$$V^*(s) = min_{a \in Ap(s)} Q^*(s, a).$$

## 2.2 Dynamic Programming

Most optimal MDP algorithms are based on dynamic programming, whose utility was first proved by a simple yet powerful algorithm named *value iteration* (Bellman, 1957). Value iteration first initializes the value function arbitrarily, for example all zero. Then, the values are updated iteratively using an operator called the *Bellman backup* (Line 7 of Algorithm 1) to create successively better approximations for each state per iteration. We define the *Bellman residual* of a state to be the absolute difference of a state value before and after a Bellman backup. Value iteration stops when the value function converges. In implementation, it is typically signaled by when the *Bellman error*,

---

**Algorithm 1** (Gauss-Seidel) Value Iteration

---

1: **Input:** an MDP $M = \langle \mathcal{S}, \mathcal{A}, Ap, T, C \rangle$, $\delta$: the threshold value
2: initialize $V$ arbitrarily
3: **while** true **do**
4:    $Bellman\_error \leftarrow 0$
5:    **for** each state $s \in S$ **do**
6:       $oldV \leftarrow V(s)$
7:       $V(s) \leftarrow min_{a \in Ap(s)} \left[ C(s,a) + \sum_{s' \in \mathcal{S}} T_a(s'|s)V(s') \right]$
8:       $Bellman\_residual(s) \leftarrow |V(s) - oldV|$
9:       $Bellman\_error \leftarrow max(Bellman\_error, Bellman\_residual(s))$
10:   **if** $Bellman\_error < \delta$ **then**
11:      **return** $V$

---

the largest Bellman residual of all states, becomes less than a pre-defined threshold, $\delta$. We call a Bellman backup a *contraction operation* (Bertsekas, 2001), if for every state, its Bellman residual never increase with the iteration number.

Value iteration converges to the optimal value function in time polynomial in $|\mathcal{S}|$ (Littman, Dean, & Kaelbling, 1995; Bonet, 2007), yet in practice it is usually inefficient, since it blindly performs backups over the state space iteratively, often introducing many unnecessary backups.

### 2.2.1 HEURISTIC SEARCH

To improve the efficiency of dynamic programming, researchers have explored various ideas from traditional heuristic-guided search, and have consistently demonstrated their usefulness for MDPs (Barto, Bradtke, & Singh, 1995; Hansen & Zilberstein, 2001; Bonet & Geffner, 2003b, 2006; McMahan et al., 2005; Smith & Simmons, 2006; Sanner et al., 2009). The basic idea of heuristic search is to consider an action only when necessary, which leads to a more conservative backup strategy. This strategy helps to save a lot of unnecessary backups.

We define a heuristic function $h : \mathcal{S} \rightarrow \mathbb{R}^+$, where $h(s)$ is an estimate of $V^*(s)$. A heuristic function $h$ is *admissible* if it never over-estimates the value of a state,

$$h(s) \leq V^*(s), \forall s \in \mathcal{S}. \tag{5}$$

We also interchangeably write an admissible heuristic function as $V_l$, to emphasize that $V_l(s)$ is a lower bound of $V^*(s)$.

**Definition** *A greedy policy $\pi$ is the best policy by one-step lookahead given the current value function, $V$:*

$$\pi(s) = argmin_{a \in Ap(s)} \left[ C(s,a) + \sum_{s' \in \mathcal{S}} T_a(s'|s)V(s') \right], \forall s \in \mathcal{S} - \mathcal{G}. \tag{6}$$

*A policy graph, $G_\pi = (\mathcal{V}, \mathcal{E})$, for an MDP with the set of states $\mathcal{S}$ and policy $\pi$ is a directed, connected graph with vertices $\mathcal{V} \subseteq \mathcal{S}$, where $s_0 \in \mathcal{V}$, and for any $s \in \mathcal{S}$, $s \in \mathcal{V}$ iff $s$ is reachable from $s_0$ under policy $\pi$. Furthermore, $\forall s, s' \in \mathcal{V}$, $\langle s, s' \rangle \in \mathcal{E}$ (the edges of the policy graph) iff $T_{\pi(s)}(s'|s) > 0$.*

Heuristic search algorithms have two main features: (1) The search is limited to states that are reachable from the initial state. Given the heuristic value, a heuristic search algorithm generates a running greedy policy, as well as its policy graph. The algorithm performs a series of heuristic searches, until all states on the greedy policy graph converge. A search typically starts from the initial state, with successor states explored in a best-first manner. Visited states have their values backed up during the search. (2) Since heuristic search algorithms do fewer backups than value iteration, they require special care to guarantee final optimality. So values of the state space have to be initialized by an admissible heuristic function. Note that value iteration can also take advantage of initial heuristic values as an informative starting point, but does not require the heuristics to be admissible to guarantee optimality.

Different heuristic search algorithms use different search strategies and therefore perform Bellman backups in different orders.

The AO* algorithm (Nilson, 1980) solves acyclic MDPs, so it is not applicable to general MDPs. LAO* (Hansen & Zilberstein, 2001) is an extension to the AO* algorithm that can handle MDPs with loops. Improved LAO* (ILAO*) (Hansen & Zilberstein, 2001) is an efficient variant of LAO*. It iteratively performs complete searches that discover a running greedy policy graph. In detail, the greedy policy graph only contains the initial state $s_0$ when a search starts. New states are added to the graph by means of *expansions* over a *frontier state* in a depth-first manner, until no more states can be added. In a state expansion, one of its greedy actions is chosen, and all the action's successor states are added into the graph. States that are not expanded yet but contain successors are called frontier states. Later, states of the greedy policy graph are backed up *only once* in the post-order when they are visited. Each search iteration performs at most $|\mathcal{S}|$ backups, but in practice this number is typically much smaller. ILAO* terminates when all states of the current greedy policy graph have a Bellman residual less than a given $\delta$.

Real-time dynamic programming (RTDP) (Barto et al., 1995) is another popular algorithm for MDPs. It interleaves dynamic programming with search through plan execution trials. An execution trial is a path that originates from $s_0$ and ends at any goal state or by a bounded-step cutoff. Each execution step simulates the result of one-step plan execution. The agent greedily picks an action $a$ of the current state $s$, and mimics the state transition to a new current state $s'$, chosen stochastically based on the transition probabilities of the action, i.e., $s' \sim T_a(s'|s)$. Dynamic programming happens when states are backed up immediately when they are visited. RTDP is good at finding a good sub-optimal policy relatively quickly. However, in order for RTDP to converge, states on the optimal policy have to be backed up sufficiently, so its convergence is usually slow. To overcome the slow convergence problem of RTDP, researchers later proposed several heuristic search variants of the algorithm.

Bonet and Geffner (2003b) introduced a smart *labeling* technique in a RTDP extension named labeled RTDP (LRTDP). They label a state $s$ *solved* if every state reachable from $s$ by applying the greedy policy is either a goal state, or is solved, or has a Bellman residual no greater than the threshold $\delta$. States that are labeled as solved no longer get backed up in any future search. Labeling helps speed up convergence as it avoids many unnecessary backups over states that have already converged. After an execution trial, LRTDP tries to label every unsolved state in the reverse order of visit. To label a state $s$, LRTDP initiates a DFS from $s_0$ and checks if all states reachable under the greedy policy rooted at $s$ are solved, and back them up, otherwise. LRTDP terminates when all states of the current policy graph are solved. Bonet and Geffner also applied the labeling technique in another algorithm called HDP (Bonet & Geffner, 2003a). HDP uses Tarjan's algorithm to find all the strongly connected component of an MDP to help label solved states and implicitly control the order in which states are backed up in a search trial.

McMahan et al. (2005) proposed another extension named bounded RTDP (BRTDP), which not only uses a lower bound heuristic of the value function $V_l$, but also an upper bound $V_u$. BRTDP has two key differences from the original RTDP algorithm. First, once BRTDP backs up a state $s$, it updates both the lower bound and the upper bound. Second, when choosing the next state $s'$, the difference of its two bounds, $V_u(s') - V_l(s')$, is also taken into consideration. More concretely, $s' \sim T_a(s'|s)[V_u(s') - V_l(s')]$, which focuses search on states that are less likely to be converged. One feature of BRTDP is its adaptive trial termination criterion, which is very helpful in practice. Smith and Simmons (2006) introduced a similar algorithm named focused RTDP (FRTDP). They define *occupancy* as an intuitive measure of the expected number of times a state is visited before execution termination. Therefore occupancy of a state indicates its relevance to a policy. Similar to BRTDP, FRTDP also keeps two bounds for a state. FRTDP uses the product of a state's occupancy and the difference of its bounds for picking the next state. Also, FRTDP assumes a discounted cost setting, so it is not immediately applicable to SSP problems.

Recently Sanner et al. (2009) described another advanced RTDP variant named Bayesian RTDP, which also uses two value bounds. The basic motivation of Bayesian RTDP is that anytime performance for sub-optimal policies is important, when finding an optimal policy can be very time-consuming. This is especially true when some sub-optimal policy performs close to an optimal one, but is much faster to generate. Its key assumption is that the true value function of a state $s$, $V^*(s)$, is uniformly distributed on the interval $[V_l(s), V_u(s)]$. Therefore, the probability density function of $V^*(s)$ is $\mathbf{1}_{v \in [V_l(s), V_u(s)]}[\frac{1}{V_u(s) - V_l(s)}]$, and $E[V^*(s)] = \frac{1}{2}[V_l(s) + V_u(s)]$. To evaluate how important it is to pick state $s'$ as the next state, it refers to the notion of *value of perfect information* (VPI), which intuitively tells the expected Q-value difference of the current state-action pair, $Q(s, a)$, with and without the knowledge of $V^*(s')$. To choose $s'$, Bayesian RTDP uses a metric that combines the BRTDP metric and the VPI value.

## 2.3 A Limitation of Previous Solvers

Value iteration backs up states iteratively based on some fixed order. Heuristic search backs up states in a dynamic, informed order, implied by when they are visited in the search. A state can be backed up in the pre-order (when it is first visited, e.g., variants of RTDP), or the post-order (when searches back track, e.g., ILAO*). None of the algorithms use an MDP's graphical structure, an intrinsic property that governs the complexity of solving a problem (Littman et al., 1995), in a way to decide the order in which states are solved.

Consider a PhD program in some Finance department. Figure 1 shows an MDP that describes the progress of a PhD student. For simplicity reasons, we omit the action nodes, the transition probabilities, and the cost functions. The goal state set is a singleton $\mathcal{G} = \{g\}$, which indicates a student gets her PhD degree. A directed edge between two states means the head state is one successor state of the tail state under at least one action. The initial state, $s_0$, describes the status of an entry-level student. She has to first pass the qualifying exam, which consists of finding a supervisor and passing an exam. Before passing the exam one can choose to work with a different supervisor (back to state $s_0$ in the figure). State $s_1$ indicates the student has found a supervisor. Then she works on her proposal, which consists of a written document and an oral exam. She has to pass both in two consecutive quarters; otherwise back to state $s_2$. After passing the proposal, at state $s_4$, she needs to defend her thesis, passing which reaches the goal state $g$.

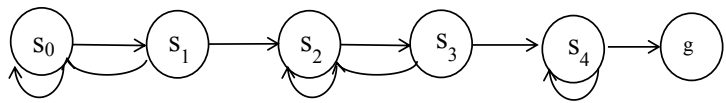

Figure 1: A simple MDP example. The action nodes, the transition probabilities, and the cost functions are omitted. The goal state set is a singleton $\mathcal{G} = \{g\}$. A directed edge between two states means the head state is one successor state of the tail state under some action.

Observing the MDP, we find the optimal order to back up states is $s_4$, then $s_2$ and $s_3$, till they converge, followed by $s_0$ and $s_1$. The reason is that the value of $s_4$ does not depend on the values of other non-goal states. Similarly, the values of $s_2$ and $s_3$ do not depend on the values of either $s_0$ or $s_1$. Value iteration as well as heuristic search algorithms do not take advantage of the graphical structure and apply this backup order, as they do not contain an "intelligent" subroutine that discovers the graphical structure, nor use this information in the dynamic programming step. The intuition of our new approaches is to discover the intrinsic complexity of solving an MDP by studying its graphical structure, which later contributes to a more intelligent backup order.

## 3. Topological Value Iteration

We now describe the topological value iteration (TVI) algorithm (Dai & Goldsmith, 2007).

First observe that the value of a state depends on the values of its successors. For example, suppose state $s_2$ is a successor state of $s_1$ under action $a$ ($T_a(s_2|s_1) > 0$). By the Bellman equations $V^*(s_1)$ is dependent on $V^*(s_2)$. In this case, we define state $s_1$ *causally depends* on state $s_2$. Note that the causal dependence relationship is transitive. We can find out all causally dependent states implicitly by building a reachability graph $G_R$ of the MDP. The set of vertices of $G_R$ equals the set of states that are reachable from $s_0$. A directed edge from vertex $s_1$ to $s_2$ means that there exists at least an action $a \in Ap(s_1)$, such that $T_a(s_2|s_1) > 0$. As the causal relationship is transitive, a directed path from state $s_1$ to $s_k$ in $G_R$ means $s_1$ is causally dependent on $s_k$, or $V^*(s_1)$ depends on $V^*(s_k)$. Also note that two vertices can be causally dependent on each other, which we call *mutual causal dependence*.

Due to causal dependence, it is usually more efficient to back up $s_2$ ahead of $s_1$. With this observation, we have the following theorem.

**Theorem 1** Optimal Backup Order *(Bertsekas, 2001): If an MDP is acyclic, then there exists an optimal backup order. By applying the optimal order, the optimal value function can be found with each state needing only one backup.*

The theorem is easy to prove and, furthermore, the optimal backup order is a topological order of the vertices in $G_R$. However, in general, MDPs contain cycles and it is common for one state to mutually causally depend on another.

If two states are mutually causally dependent, the best order to back up them is unclear. On the other hand, if neither state is causally dependent on the other, the order of backup does not matter. Finally, if one state is causally dependent on the other (and not vice versa), it is better to order the backups so that the state which is causally dependent is updated later. To apply this idea we then group together states that are mutually causally dependent and make them a *meta-state*. We make a new directed graph $G_M$ where a directed edge between two meta-states $\mathcal{X}$ and $\mathcal{Y}$ exists if and only if there exists two states $s_1$ and $s_2$ and an action $a \in Ap(s_1)$ such that $s_1 \in \mathcal{X}$, $s_2 \in \mathcal{Y}$ and $T_a(s_2|s_1) > 0$. It is clear that $G_M$ is acyclic, otherwise all states on such a cycle are mutually causally dependent, and by our construction rule they should belong to the same meta-state. In this case, we can back up states in $G_M$ in their topological order. By Theorem 1, each such state only requires one *meta-backup*. It is called a meta-backup since a meta-state may contain multiple states. To perform a meta-backup, we can apply any dynamic programming algorithm, such as value iteration, on all states belonging to the corresponding meta-state.

The pseudo-code of TVI is shown in Algorithm 2. We first apply Kosaraju's algorithm (Cormen, Leiserson, Rivest, & Stein, 2001) to find the set of strongly connected components (SCCs, or meta-states) in the causality graph $G_R$, and its topological order. ($id[s]$ indicates the topological order of the SCC that state $s$ belongs to.) It is based on the fact that by reversing all the edges in $G_R$, the resulting graph, $G'_R$, has the same strongly connected components as the original. From using that, we can get the SCCs by doing a forward traversal to find an ordering of vertices, followed by a traversal of the reverse of the graph in the order generated by the first traversal. Kosaraju's algorithm is efficient, as its time complexity is linear in the number of states. When the state space is large, running the algorithm leads to unavoidable yet acceptable overhead. In many cases the overhead is well compensated by the computational gain. We then use value iteration to solve each SCC $\mathcal{C}$ (as a meta-backup) in its topological order.

---

**Algorithm 2** Topological Value Iteration

---

1: **Input:** an MDP $M = \langle \mathcal{S}, \mathcal{A}, Ap, T, C \rangle$, $\delta$: the threshold value
2: **SCC**$(M)$
3: **for** $i \leftarrow 1$ to $cpntnum$ **do**
4:     $\mathcal{S}' \leftarrow$ the set of states $s$ where $id[s] = i$
5:     $M' \leftarrow \langle \mathcal{S}', \mathcal{A}, Ap, T, C \rangle$
6:     **VI**$(M', \delta)$
7:
8: **Function** $SCC(M)$
9: construct $G_R$ of $M$
10: construct a graph $G'_R$ which reverses the head and tail vertices of every edge in $G_R$
11: {call Kosaraju's algorithm (Cormen et al., 2001). It inputs $G_R$ and $G'_R$ and outputs $cpntnum$, the total number of SCCs, and $id : \mathcal{S} \rightarrow [1, cpntnum]$, the id of the SCC each state belongs to, by topological order.}
12: **return** $(cpntnum, id)$

---

## 3.1 Convergence

When the Bellman operator is a *contraction operation* (Bertsekas, 2001), we have:

**Theorem 2** *Topological Value Iteration is guaranteed to converge to a value function with a Bellman error that is no greater than $\delta$.*

**Proof** We first prove that TVI is guaranteed to terminate in finite time. Since each MDP contains a finite number of states, it contains a finite number of connected components. In solving each of these components, TVI uses value iteration. Because value iteration is guaranteed to converge in finite time (given a finite $\delta$), TVI, which is essentially a finite number of value iterations, terminates in finite time.

We then prove TVI is guaranteed to converge to an optimal value function with Bellman error at most $\delta$. We prove by induction.

First, if an MDP contains only one SCC, then TVI coincides with VI, an optimal algorithm. By the contraction property of Bellman backups, when VI converges, the Bellman error of the state space is at most $\delta$.

Now, consider the case where an MDP contains multiple SCCs. At any point, TVI is working on one component $\mathcal{C}$. We know that the optimal value of every state $s \in \mathcal{C}$, $V^*(s)$, depends only on the optimal values of the states that are descendants of $s$. We also know that any descendant $s'$ of $s$ must belong either to $\mathcal{C}$, or a component $\mathcal{C}'$ that is topologically no later than $\mathcal{C}$. This means either its value is computed by VI in the same batch as $s$ ($s' \in \mathcal{C}$), or state $s'$ is already converged ($s' \in \mathcal{C}'$). In the latter case, its value is a convex combination of states having error at most $\delta$. Inside each maximization operation of an Bellman equation is an affine combination of values with a total weight of 1, which leads to an overall convex combination error of no more than $\delta$. Therefore, when VI finishes solving $\mathcal{C}$, the value of $s$ must converge with Bellman residual at most $\delta$. Also note that the values of all states that belong to a component that is earlier than $\mathcal{C}$ does not depend on those of states in component $\mathcal{C}$. As a result, after component $\mathcal{C}$ converges, the Bellman residual of states in those components remain unchanged and thus are at most $\delta$. Combining the results we conclude that when TVI terminates, the Bellman residuals of all states are at most $\delta$. This means the Bellman error of the state space is at most $\delta$.

From the high-level perspective, TVI decomposes an MDP into sub-problems and finds the value of the state space in a batch manner, component by component. When a component is converged, all its states will be safely treated as sink states, as their values do not depend on values of states belonging to later components.

## 3.2 Implementation

We made two optimizations in implementing TVI. The first one is an uninformed reachability analysis. TVI does not depend on any initial state information. However, once given that information, TVI is able to mark the reachable components and later ignore the unreachable ones in the dynamic programming step. The reachable state space can be found by a depth-first search starting from $s_0$, with an overhead that is linear in $|\mathcal{S}|$ and $|\mathcal{A}|$. It is extremely useful when only a small portion of the state space is reachable (e.g., most domains from the International Planning Competition 2006, see Bonet, 2006).

The second optimization is to use heuristic values $V_l(\cdot)$ as a starting point. We used the $h_{min}$ (Bonet & Geffner, 2003b), an admissible heuristic:

$$
\begin{aligned}
h_{min}(s) &= 0 \quad \text{if } s \in \mathcal{G}, \text{ else} \\
h_{min}(s) &= \min_{a \in Ap(s)} \left[ C(s,a) + min_{s':T_a(s'|s)>0} h_{min}(s') \right].
\end{aligned}
\tag{7}
$$

To implement it, we first construct a new deterministic problem. For each action and successor pair of the original MDP, we add to the new problem a deterministic action with the same cost and the same, deterministic successor. We then solve this new problem by a single, backward, breadth-first search from the set of goal states. Values of the deterministic problem are $h_{min}$.

### 3.3 Experiments

We address the following questions in our experiments: (1) How does TVI compare with VI and heuristic search algorithms on MDPs that contain multiple SCCs? (2) What are the most favorable problem features for TVI?

We compared TVI with several other optimal algorithms, including VI (Bellman, 1957), ILAO* (Hansen & Zilberstein, 2001), LRTDP (Bonet & Geffner, 2003b), BRTDP (McMahan et al., 2005), Bayesian RTDP (Sanner et al., 2009) (BaRTDP), and HDP (Bonet & Geffner, 2003a)[1]. We used the fully optimized C code of ILAO* provided by Eric A. Hansen and additionally implemented the rest of the algorithms over the same framework. We performed all experiments on a 2.5GHz Dual-Core AMD Opteron(tm) Processor with 2GB memory. Recall that BRTDP and BaRTDP use upper bounds. We used upper bounds as described in Section 4.2. We used $\alpha = 2 \times 10^{-6}$ and $\tau = 10$ for BRTDP and BaRTDP.[2] For BaRTDP, we used the probabilistic termination condition in Algorithm 3 of Sanner et al. (2009). [3]

We compared all algorithms on *running time*, time between an algorithm starts solving a problem until generating a policy with a Bellman error of at most $\delta(= 10^{-6})$. We terminated an algorithm if it did not find such a policy within five minutes. Note that there are other performance measures such as anytime performance (the original motivation of BaRTDP) and space consumption, but the main motivation of TVI is to decrease convergence time. We expect TVI to have a very steep anytime performance curve, because it postpones backing up the initial state till it starts working on the SCC where the initial state belongs to. Space, on the other hand, is less interesting because in-memory MDPs algorithms requires that the MDP model stored in the main memory before dynamic programming can apply. Therefore, they all share the same space limit. For work on overcoming space limitation, see, for example the work of Dai et al. (2008, 2009a).

We tested all algorithms on a set of artificially-generated "layered" MDPs. For each such MDP of state size $|\mathcal{S}|$, we partition the state space evenly into a number $n_l$ of *layers*, labeled by integers $1, \ldots, n_l$. We allow states in higher numbered layers to be the successors of states in lower numbered layers, but not vice versa, so each state $s$ only has a limited set of allowable successor states, named $succ(s)$. A layered MDP is parameterized by two other variables: the number of actions per state, $n_a$, and the maximum number of successor states per action, $n_s$. When generating the transition function of a state-action pair $(s, a)$, we draw an integer $k$ uniformly from $[1, n_s]$. Then $k$ distinct successors are uniformly sampled from $succ(s)$ with random transition probabilities. We pick one state from layer $n_l$ as the only goal state. One property of a layered MDP is that it contains at least $n_l$ connected components.

---

1. Notice that this comparison is somewhat unfair to TVI, since heuristic search algorithms may not expand portions of the state space, if their sub-optimality can be proved. Still, we make this comparison to understand the practical benefits of TVI v.s. all other known optimal MDP algorithms

2. $\alpha$ is the termination threshold of BRTDP (it terminates when $v_u(\mathbf{s_0}) - V_l(\mathbf{s_0}) < \alpha$). $\tau$ indicates the stopping condition of each heuristic search trial. For more detailed discussions on the two parameters, please refer to the work of McMahanet al. (2005). We carefully tuned these parameters.

3. This termination condition may result in sub-optimal policies, so the reported times of BaRTDP in this paper are lower bounds. Note that BaRTDP mainly aims at improving the anytime performance of RTDP, which is *orthogonal* to convergence time. We report its convergence speed for thorough investigation purposes.

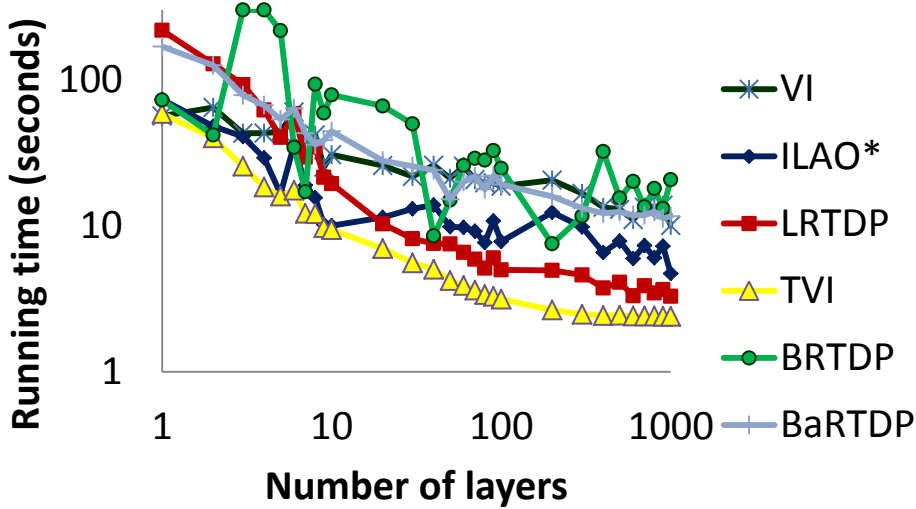

Figure 2: Running times of algorithms with different number of layers $n_l$ on random layered MDPs with $|\mathcal{S}| = 50000$, $n_a = 10$, and $n_s = 10$. Note that the two coordinates are both log-scaled. When $n_l > 10$ TVI not only outperforms VI, but also other state-of-the-art heuristic search algorithms.

There are several planning domains that lead to multi-layered MDPs. An example is the game Bejeweled, or any game with difficulty levels: each level is at least one layer. Or consider a chess variant without pawn promotions, played against a stochastic opponent. Each set of pieces that could appear on the board together leads to at least one strongly connected component. But we know of no multi-layered standard MDP benchmarks. Therefore, we compare, in this section, on artificial problems to study TVI's performance across controlled parameters, such as $n_l$ and $|\mathcal{S}|$. Next section contains more comprehensive experiments on benchmark problems.

We generated problems with different parameter configurations and ran all algorithms on the same set of problems. The running times, if the process converged within the cut-off, are reported in Figures 2 and 3. Each element of the table represents the median convergence time of running 10 MDPs with the same configuration.[4] Note that varying $|\mathcal{S}|$, $n_l$, $n_a$, and $n_s$ yields many MDP configurations. We tried more combinations than the representative ones reported. We found HDP much slower than the other algorithms, so did not include its performance.

For the first experiment, we fixed $|\mathcal{S}|$ to be 50,000 and varied $n_l$ from 1 to 1,000. Observing Figure 2 we first find that, when there is only one layer, the performance of TVI is slightly worse than VI, as such an MDP probably contains an SCC that contains the majority of the state space, which defeats the benefit of TVI. But TVI consistently outperforms VI if $n_l > 1$. When $n_l \leq 10$, TVI equals or beats ILAO*, the fastest heuristic search algorithm for this set of problems. When $n_l > 10$, TVI outperforms all the other algorithms in all cases by a visible margin. Also note that, as the number of layers increases the running times of all algorithms decrease. This is because the MDPs become more structured, therefore simpler to solve. The running time of TVI decreases second fastest to that of LRTDP. LRTDP is very slow when $n_l = 1$ and its running time drops dramatically when $n_l$ increases from 1 to 20. As TVI spends nearly constant time in generating the topological order of the SCCs, its fast convergence is mainly due to the fact that VI is much more efficient in solving many small (and roughly equal-sized) problems than a large problem whose size is the same as the sum of the small ones. This experiment shows TVI is good at solving MDPs with many SCCs.

---

4. We picked median instead of mean just to avoid an unexpected hard problem, which takes a long time to solve, thereby dominating the performance.

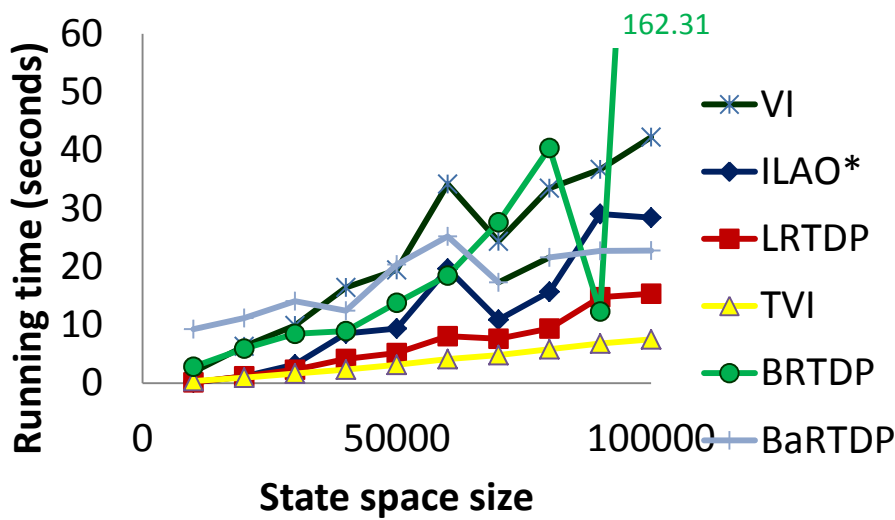

Figure 3: Running times of algorithms with different state space size $|\mathcal{S}|$ with fixed $n_l = 100$, $n_a = 10$, and $n_s = 10$. TVI not only outperforms VI, but also other state-of-the-art heuristic search algorithms. The relative performance of TVI improves as $|\mathcal{S}|$ increases.

For the second experiment, we fixed $n_l$ to be 100 and varied $|\mathcal{S}|$ from 10,000 to 100,000. We find that, when the state space is 10,000 TVI outperforms VI, BRTDP and BaRTDP, but slightly underperforms ILAO* and LRTDP. However, as the problem size grows TVI soon takes the lead. It outperforms all the other algorithms when the state space is 20,000 or larger. When the state space grows to 100,000, TVI solves a problem 6 times as fast as VI, 4 times as fast as ILAO*, 2 times as fast as LRTDP, 21 times as fast as BRTDP, and 3 times as fast as BaRTDP. This experiment shows that TVI is even more efficient when the problem space is larger.

## 4. Focused Topological Value Iteration

Topological value iteration improves the performance of value iteration most significantly when an MDP has many equal-sized strongly connected components. However, we also observe that many MDPs do not have evenly distributed connected components. This is due to the following reason: a state can have many actions, most of which are sub-optimal. These sub-optimal actions, although not part of an optimal policy, may lead to connectivity between a lot of states. For example, domains like Blocksworld have reversible actions. Due to these actions most states are mutually causally dependent. As a result, states connected by reversible actions end up forming a large connected component, making TVI slow.

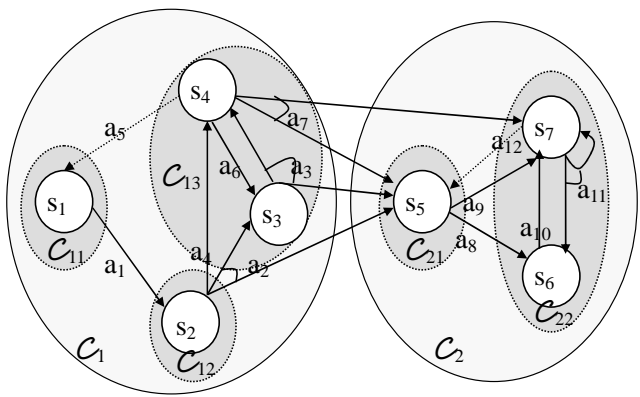

Figure 4: The graphical representation of an MDP and its set of strongly connected components (before and after the knowledge of some sub-optimal actions). Arcs represent probabilistic transitions, e.g., $a_7$ has two probabilistic successors – $s_5$ and $s_7$.

On the other hand, heuristic search is a powerful solution technique, which successfully concentrates computation, in the form of backups, on states and transitions that are more likely to be part of an optimal policy. However, heuristic search uses the same backup strategy on all problems, thus missing out on the potential savings from knowing the graphical structure information.

If we knew about the existence of an action in the optimal policy, we could eliminate the rest actions for its outgoing state, thus breaking some connectivity. Of course, such information is never available. However, with a little help from heuristic search, we can eliminate sub-optimal actions from a problem leading to a reduced connectivity and hopefully, smaller sizes of strongly connected components.

Figure 4 shows the graphical representation of a part of one simple MDP that has 7 states and 12 actions. In the figure, successors of probabilistic actions are connected by an arc. For simplicity, transition probabilities $T_a$, costs $C(s, a)$, initial state and goal states are omitted. Using TVI, we can divide the MDP into two SCCs $\mathcal{C}_1$ and $\mathcal{C}_2$. However, suppose we are given some additional information that $a_5$ and $a_{12}$ are sub-optimal. Based on the remaining actions, $\mathcal{C}_1$ and $\mathcal{C}_2$ can be sub-divided into three and two smaller components respectively (as shown in the figure). Dynamic programming will greatly benefit from the new graphical structure, since solving smaller components can be much easier than a large one.

## 4.1 The FTVI Algorithm

The key insight of our novel algorithm is to break the big components into smaller parts, by removing actions that can be proven to be suboptimal for the current problem at hand. This exploits the knowledge of the current initial state and goal, which TVI mostly ignores. We call our new algorithm *focused topological value iteration* (FTVI) (Dai et al., 2009b). The pseudo-code is shown in Algorithm 3.

At its core, FTVI makes use of the *action elimination* theorem, which states:

**Theorem 3** Action Elimination *(Bertsekas, 2001): If a lower bound of $Q^*(s, a)$ is greater than an upper bound of $V^*(s)$ then action $a$ cannot be an optimal action for state $s$.*

This gives us a template to eliminate actions, except that we need to compute a lower bound for $Q^*$ and an upper bound for $V^*$. FTVI keeps two bounds of $V^*$ simultaneously: the lower bound $V_l(\cdot)$ and the upper bound $V_u(\cdot)$. $V_l(\cdot)$ is initialized via the admissible heuristic. We note two properties of $V_l$: (1) $Q_l(s, a)$ computed by a one-step lookahead given the current lower bound value $V_l(\cdot)$ (Line 30, Algorithm 3) is a lower bound of $Q^*(s, a)$, and (2) all the $V$ values remain lower bounds throughout the algorithm execution process, if they were initialized by an admissible heuristic. So, this lets us easily compute a lower bound of $Q^*$, which also improves as more backups are performed.

Similar properties hold for $V_u$, the upper bound of $V^*$, i.e., if we initialize $V_u$ by an upper bound and perform backups based on $V_u$ then each successive value estimate remains an upper bound. The later implementation section lists our exact procedure to compute the lower and upper bounds in a domain-independent manner. We note that to employ action elimination we can use any lower and upper bounds, so if a domain has informative, domain-dependent bounds available, that can be easily plugged into FTVI.

FTVI contains two sequential steps. In the first step, which we call the *search* step, FTVI performs a small number of heuristic searches similar to ILAO*, i.e., backs up a state at most once per iteration. This makes the searches in FTVI fast, but still useful enough to eliminate sub-optimal actions. There are two main differences in common heuristic search and the search phase of FTVI. First, in each backup, we update the upper bound in the same manner as the lower bound. This is reminiscent of backups in BRTDP (McMahan et al., 2005). Second, we also check and eliminate sub-optimal actions using action elimination (Lines 30–32).

In the second step, the *computation* step, FTVI generates a directed graph $G_{SR}$ in the same manner as TVI generates $G_R$, but only based on the remaining actions. More concretely, a directed edge from vertex $s_1$ to $s_2$ exists if there is an *uneliminated* action $a$ such that $T_a(s_2|s_1) > 0$. It is easy to see that the graph $G_{SR}$ generated is always a sub-graph of $G_R$. FTVI then finds all connected components of $G_{SR}$, their topological order, and solves each component sequentially in the topological order.

We can state the following theorem for FTVI.

**Theorem 4** *FTVI is guaranteed to converge to the optimal value function.*

The correctness of the theorem is based on two facts: (1) action elimination preserves soundness, and (2) TVI is an optimal planning algorithm (Theorem 2).

### 4.2 Implementation

There are several interesting questions to answer in implementation. How to calculate the initial upper and lower bounds? How many search iterations do we need to perform in the search step? Is it possible that FTVI converges in the search step? What if there still remains a large component even after action elimination?

We used the same lower bound $V_l$ as in TVI (see Section 3.2). For the upper bound, we started with a simple upper bound:

---

**Algorithm 3** Focused Topological Value Iteration

---

1: **Input:** an MDP $\langle \mathcal{S}, \mathcal{A}, Ap, T, C \rangle$, $x$: the number of search iterations in a batch, $y$: the lower bound of the percentage of change in the initial state value for a new batch of search iterations, $\delta$: the threshold value

2: {step 1: search}

3: **while** true **do**

4:   $old\_value \leftarrow V_l(s_0)$

5:   **for** $iter \leftarrow 1$ to $x$ **do**

6:     $Bellman\_error \leftarrow 0$

7:     **for** every state $s$ **do**

8:       mark every state as unvisited

9:     $s \leftarrow s_0$

10:    $Search(s)$

11:    **if** $Bellman\_error < \delta$ **then** {The value function converges}

12:      **return** $V_l$

13:    **if** $old\_value/V_l(s_0) > (100 - y)\%$ **then**

14:      **break**

15:

16: {step 2: computation}

17: $M \leftarrow \langle \mathcal{S}, \mathcal{A}, Ap, T, C \rangle$

18: **TVI**$(M, \delta)$ {by applying the backup operator with action elimination}

19:

20: **Function** $Search(s)$

21: **if** $s \notin \mathcal{G}$ **then**

22:   mark $s$ as visited

23:   $a \leftarrow argmin_a Q(s, a)$

24:   **for** every unvisited successor $s'$ of action $a$ **do**

25:     $Search(s')$

26:   $Bellman\_error \leftarrow max(Bellman\_error, Back - up(s))$

27:

28: **Function** $Back - up(s)$

29: **for** each action $a$ **do**

30:   $Q(s, a) \leftarrow C(s, a) + \sum_{s' \in \mathcal{S}} T_{a'}(s'|s) V_l(s')$

31:   **if** $Q_l(s, a) > V_u(s)$ **then**

32:     eliminate $a$ from $Ap(s)$

33: $oldV_l \leftarrow V_l(s)$

34: $V_l(s) \leftarrow min_{a \in Ap(s)} Q(s, a)$

35: $V_u(s) \leftarrow min_{a \in Ap(s)} [C(s, a) + \sum_{s' \in \mathcal{S}} T_{a'}(s'|s) V_u(s')]$

36: **return** $|V_l(s) - oldV_l|$

---

$$V_u(s) = 0 \quad \text{if } s \in \mathcal{G}, \text{ else} \quad V_u(s) = \infty. \qquad (8)$$

This initialization gives us a global yet very loose upper bound. To improve its tightness, we performed a backward best-first search from the set of goal states. States visited have their $V_u$ values updated as in Algorithm 3, Line 35. We can iteratively get tighter and tighter bounds when more backward searches are performed.

The time spent on search can have a significant impact on FTVI. Very few search iterations might not eliminate enough sub-optimal actions. However, too many search iterations will turn

FTVI into a heuristic search algorithm and trade off the advantage of FTVI. We did a control experiment by varying the total number of heuristic search trials on two problems. Figure 5 shows that the performance on a Wet-floor problem matches our hypothesis perfectly. For the Drive problem, the number of search trials does not affect the convergence speed too much, but too many search trials turn out to be harmful.

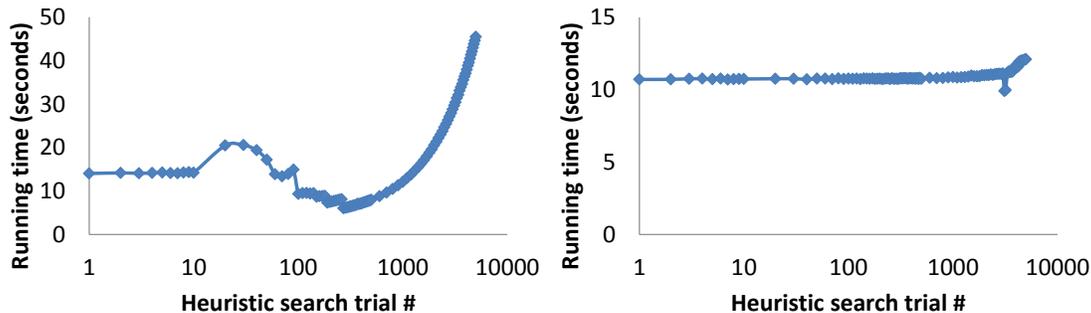

Figure 5: Running times of FTVI with different number of initial search trials on (left) a Wet-floor problem and (right) a Drive problem. Too few trials are sometimes less helpful for eliminating enough sup-optimal actions, and too many trials are harmful.

Considering the tradeoff, we let the algorithm automatically determine the number of search iterations. FTVI incrementally performs a batch of $x$ search iterations. After the batch, it computes the amount of change to the $V_i(s_0)$ value. If the change is greater than $y\%$, a new batch of search is performed. Otherwise, the search phase is considered complete. In our implementation, we use $x = 100$, and $y = 3$.

An interesting case occurs when the optimal value is found during the search step. Although FTVI performs a limited number of search iterations, it is possible that a problem is optimally solved within the search step. It is helpful to keep track of optimality information during the search step, so that FTVI can potentially skip some unnecessary search iterations and the entire computation step. To do this, we only need to maintain a Bellman error of the current search iteration, and terminate FTVI if the error is smaller than the threshold (Lines 11–12). In our experiment, we find this simple optimization to be extremely helpful in promoting the performance of FTVI.

Sometimes there are cases where $G_{SR}$ still contains large connected components. This can be caused by two reasons (1) An optimal policy indeed has large components, or (2) the connectivity caused by many suboptimal actions is not successfully eliminated by search. To try to further decompose these large components, we let FTVI perform additional *intra-component heuristic searches*. An intra-component heuristic search takes place only inside a particular component. Its purpose is to find new, sub-optimal actions, which might help decompose the component. Given a component $\mathcal{C}$ of $G_{SR}$, we define $Source_{\mathcal{C}}$ to be the set of states where none of its incoming transitions are from states in $\mathcal{C}$. In other words, states in $Source_C$ are the incoming bridge states between $\mathcal{C}$ and rest of the MDP. An intra-component heuristic search of $\mathcal{C}$ originates from a state in $Source_{\mathcal{C}}$. A search branch terminates when a state outside $\mathcal{C}$ is encountered.

We did some experiments and compared the performance of FTVI with and without additional intra-component search on problems from four domains, namely Wet-floor (Bonet & Geffner, 2006),

Single-arm pendulum (Wingate & Seppi, 2005), Drive, and Elevator (Bonet, 2006). Our results show that additional intra-component search only provided limited gains in Wet-floor problems, in which it helped decrease the size of the largest components by approximately 50% on average, and sped up the convergence by 10% at best. However, intra-component search turned out to be harmful for the other domains, as it did not provide any new graphical information (no smaller components were generated). On the contrary, the search itself introduced a lot of unnecessary overhead. So we used the version that does not perform additional intra-component search throughout the rest of the experiments.

## 4.3 Experiments

We address the following two questions in our experiments: (1) How does FTVI compare with other algorithms on a broad range of domain problems? (2) What are the specific kind of domains on which FTVI should be preferred over heuristic search?

We implemented FTVI on the same framework as in Section 3.3, and used the same cut-off time of 5 minutes for each algorithm per problem. To investigate the helpfulness of action elimination, we also implemented a VI variant that applies action elimination in backups. We used the same threshold value $\delta = 10^{-6}$, and ran BRTDP and BaRTDP on the same upper bound as FTVI.

### 4.3.1 RELATIVE SPEED OF FTVI

| Problem | VI | VI (w/ a.e.) | ILAO* | LRTDP | BRTDP | BaRTDP | TVI | FTVI |
|---|---|---|---|---|---|---|---|---|
| MCar100 | 1.40 | 0.74 | 1.91 | 1.23 | 2.81 | 63.55 (*) | 0.68 | **0.22** |
| MCar300 | 26.12 | 13.40 | 11.91 | 229.70 | 117.23 | 180.64 (*) | 23.22 | **2.35** |
| MCar700 | 278.16 | 124.34 | 101.65 | - | 216.01 | 262.92 (*) | 233.98 | **13.06** |
| SAP100 | 2.30 | 1.06 | 1.81 | 2.58 | 9.39 | 111.59 (*) | 2.37 | **0.17** |
| SAP300 | 42.61 | 19.90 | 32.40 | - | - | - | 44.2 | **2.96** |
| SAP500 | 174.71 | 77.99 | 131.17 | - | - | - | - | **9.56** |
| WF200 | 19.95 | 13.71 | 11.22 | - | 22.08 | 1.99 (*) | 20.58 | **8.81** |
| WF400 | 105.79 | 98.97 | **73.88** | - | 97.73 | 103.87 (*) | 100.78 | 74.24 |
| DAP10 | 0.77 | 0.67 | 1.01 | 51.45 | 3.04 | 222.33 (*) | 0.75 | **0.59** |
| DAP20 | 21.41 | 17.62 | 32.68 | - | 144.12 | - | 21.95 | **17.49** |
| Drive | 2.00 | 1.39 | 1.60 | **0.69** | 7.85 | 4.17 (*) | 1.23 | 1.07 |
| Drive | 20.58 | 14.20 | 96.09 | 273.37 | 163.91 | 4.17 (*) | 13.03 | **10.63** |
| Drive | - | - | - | - | - | 3.94 (*) | 74.70 | **41.93** |
| Elevator (IPPC p13) | - | - | 227.53 | - | - | - | 58.46 | **54.11** |
| Elevator (IPPC p15) | 236.91 | 133.80 | 27.35 | - | - | - | 14.59 | **12.11** |
| Tireworld (IPPC p5) | 33.88 | 16.46 | **0.00** | 0.14 | 0.01 | 0.03 | 2.26 | **0.00** |
| Tireworld (IPPC p6) | 47.88 | 23.04 | **0.00** | 0.16 | 0.01 | 0.04 | 48.81 | **0.00** |
| Blocksworld (IPPC p4) | 17.69 | 17.69 | **0.02** | 0.26 | 1.93 | - | 54.35 | **0.02** |
| Blocksworld (IPPC p5) | 14.19 | 14.19 | **0.00** | 0.11 | 0.66 | - | 54.34 | **0.00** |

Table 1: Total running times of the different algorithms on problems in various domains. FTVI outperforms all algorithms by vast margins. (Fastest times are bolded. '-' in Time means that the algorithm failed to solve the problem within 5 minutes. The *'s mean the algorithm terminated with sub-optimal solutions.)

| Problem | Reachable $|\mathcal{S}|$ | TVI | | FTVI | | | |
|---|---|---|---|---|---|---|---|
| | | BC size | $Time$ | BC size | $T_{search}$ | $T_{gen}$ | $Time$ |
| MCar100 | 10,000 | 7,799 | 0.68 | 1 | 0.20 | 0.01 | **0.22** |
| MCar300 | 90,000 | 71,751 | 23.22 | 1 | 2.22 | 0.13 | **2.35** |
| MCar700 | 490,000 | 390,191 | 233.98 | 1 | 12.29 | 0.76 | **13.06** |
| SAP100 | 10,000 | 9,999 | 2.37 | n/a | 0.17 | n/a | **0.17** |
| SAP300 | 90,000 | 89,999 | 44.2 | n/a | 2.96 | n/a | **2.96** |
| SAP500 | 250,000 | - | - | n/a | 9.56 | n/a | **9.56** |
| WF200 | 40,000 | 39,999 | 20.58 | 15,039 | 3.30 | 0.12 | **8.81** |
| WF400 | 160,000 | 159,999 | 100.78 | 141,671 | 14.27 | 0.36 | 74.24 |
| DAP10 | 10,000 | 9,454 | 0.75 | n/a | 0.59 | n/a | **0.59** |
| DAP20 | 160,000 | 150,489 | 21.95 | n/a | 17.49 | n/a | **17.49** |
| Drive | 4,563 | 4,560 | 1.23 | 4,560 | 0.11 | 0.02 | 1.07 |
| Drive | 29,403 | 29,400 | 13.03 | 29,400 | 0.15 | 0.15 | **10.63** |
| Drive | 75,840 | 75,840 | 74.70 | 75,840 | 0.18 | 0.40 | **41.93** |
| Elevator (IPPC p13) | 539,136 | 1,053 | 58.46 | 1,053 | 0.01 | 1.73 | **54.11** |
| Elevator (IPPC p15) | 539,136 | 1,053 | 14.59 | 1,053 | 0.01 | 1.60 | **12.11** |
| Tireworld (IPPC p5) | 671,687 | 23 | 2.26 | n/a | 0.00 | n/a | **0.00** |
| Tireworld (IPPC p6) | 724,933 | 618,448 | 48.81 | n/a | 0.00 | n/a | **0.00** |
| Blocksworld (IPPC p4) | 103,121 | 103,104 | 54.35 | n/a | 0.02 | n/a | **0.02** |
| Blocksworld (IPPC p5) | 103,121 | 103,204 | 54.34 | n/a | 0.00 | n/a | **0.00** |

Table 2: Detailed performance statistics for TVI and FTVI. (BC size means the size of the biggest connected component. 'n/a' means FTVI converged in the search step and skipped the computation step. All running times are in seconds. $T_{search}$ represents the time used by the search step, and $T_{gen}$ the time spent in generating the graphical structure. Fastest times are bolded. '-' in Time means that the algorithm failed to solve the problem within 5 minutes.)

We evaluated the various algorithms on problems from eight domains — Mountain Car, Single and Double Arm Pendulum (Wingate & Seppi, 2005), Wet-floor (Bonet & Geffner, 2006)[5], and four domains from International Planning Competition 2006 — Drive, Elevators, TireWorld and Blocksworld. A mountain car problem usually has many source states.[6] We chose each source state as an initial state, and averaged the statistics per problem. Table 1 lists the running times for the various algorithms. For FTVI, we additionally report (in Table 2) the time used by the searches ($T_{search}$), and the time spent in generating the graphical structure ($T_{gen}$), if a problem is not solved during the search phase, where the leftover is the time spent in solving the SCCs. We also compared the size of the biggest component (BC size) generated by TVI and FTVI.

Overall we find that FTVI outperforms the other five algorithms on most of these domains. FTVI outperforms TVI in all domains. Notice that on the MCar problems, FTVI establishes very favorable graphical structures (strongly connected components of size one) during the search step.[7] This graphical structure makes the second step of FTVI trivial. But TVI has to solve much bigger components, so it runs much slower. For the Drive domain, even if it does not find a more informed graphical structure, the advanced backup with action elimination enables FTVI converge faster.

---

5. Note that we used the probability of wet cells, $p = 0.5$.

6. A source state is a state with no incoming transitions.

7. If we allow FTVI to perform the computation step as opposed to stop at the search step when a problem is solved, it will find similar structures in the Tireworld and Blocksworld problems.

FTVI outperforms heuristic search algorithms most significantly in domains such as MCar, SAP and Drive. It is faster than ILAO* by an order of magnitude. This shows the extreme effectiveness of FTVI's decomposing a problem into small sub-problems using advanced graphical information and solving these sub-problems sequentially. The three RTDP algorithms are not competitive with the other algorithms in these domains, and fail to return a solution by the cutoff time for many problems. FTVI shows limited speedup against heuristic search in domains such as Wet-floor, DAP, and Elevator. FTVI is on par with ILAO*, and vastly outperforms TVI in Tireworld and Blocksworld domains, as it converges within the search step. The convergence speed of value iteration is typically slow, as it backs up states iteratively by a fixed order. Adding action elimination to Bellman backups increases the convergence speed of VI up to two times, especially in the Mountain Car, Single Arm Pendulum, and Elevator domains, but its convergence speed is usually at least one magnitude slower than those of FTVI.

### 4.3.2 FACTORS DETERMINING PERFORMANCE

We have shown that FTVI is faster than heuristic search algorithms in many domains, but its relative speedup is domain-dependent. Can we find any domain features that are particularly beneficial for FTVI or worse for heuristic search algorithms? In this evaluation we performed control experiments by varying the domains across different features and study the effect on planning time of various algorithms.

We make an initial prediction of three features.

1. The number of goals in the domain: If the number of goal states is small, search may take a long time before it discovers a path to a goal. Therefore, many sub-optimal policies might be evaluated by a heuristic search algorithm.

2. Search depth from the initial state to a goal state: This depth is a lower bound of the length of an execution trial and also of the *size* of any policy graph. A greater depth implies more search steps per iteration, which might make evaluating a policy time-consuming.

3. Heuristic informativeness: The performance of a heuristic search algorithm depends a lot on the quality of the initial heuristic function. We expect the win from FTVI to increase when heuristic is less informed.

*The Number of Goals.* As far as we know, there is no suitable domain where we can specify the total number of goal states arbitrarily, so we used an artificial domain. In this domain each state has two applicable actions, and each action has at most two random successors. We tested all algorithms on domains of two sizes, 10,000 (Figure 6(left)) and 50,000 (Figure 6(right)). For each problem size, we fixed the shortest goal distance but varied the number of goal states, $|\mathcal{G}|$. More concretely, after generating the state transitions, we performed a BFS from the initial state, and randomly picked goal states on a same search depth. For each $|\mathcal{G}|$ value, we generated 10 problems, and reported the median running time of four algorithms (LRTDP and BaRTDP were slow in this domain). We observe that all algorithms take more time to solve a problem with a smaller number of goal states than with a larger number. However, beyond a point ($|\mathcal{G}| > 20$ in our experiments), the running times become stable. FTVI runs only marginally slower when $|\mathcal{G}|$ is small, suggesting that its performance is less dependent on the number of goal states. BRTDP is the second best in handling small goal sets, and it runs nearly as fast as FTVI when the goal set is large. Even though TVI runs the slowest among the four algorithms, its performance shows less severe dependence on the number of goal states. It runs almost as fast as ILAO* when the goal set size is 1. In contrast, ILAO* runs twice as fast as TVI when the goal set size is greater than 20.

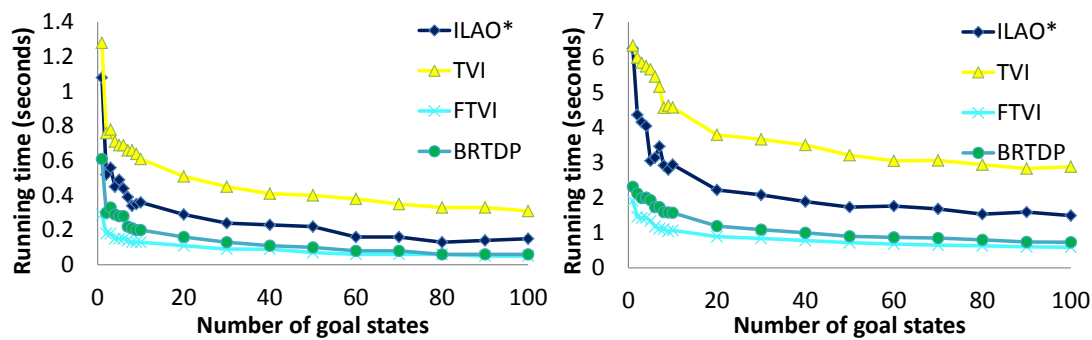

Figure 6: Running times of algorithms with different number of goal states and problem size (left) $|S| = 10,000$ (right) $|S| = 50,000$ in random MDPs. FTVI and TVI slow down the least significantly when the number of goal states is small.

*Search Depth.* In this experiment, we studied how the search depth of a goal from the initial state influences the performance of various algorithms. We chose a Mountain car problem and a Single-arm pendulum problem. We randomly picked 100 initial states from the state space[8] and measured the shallowest search depth, or, the shortest distance, $d$, to a goal state. The running times in Figure 7 are ordered by $d$. BaRTDP does not terminate with an optimal policy for many instances, so its performance is not shown. BRTDP has the biggest variance so its performance is not included for clarity purposes.

As we can see, FTVI is the fastest algorithm in this suite of experiments. It converges very quickly for all initial states (usually around one or two seconds on Mcar300, and less than 10 seconds on SAP300). TVI's performance is unaffected by the search depth, which is expected, since it is a variant of value iteration and has no search component. In the MCar300 problem, we do not find strong evidence that the running time of any algorithm depends on the search depth. FTVI runs an order of magnitude faster than TVI, ILAO*, and BRTDP and two orders of magnitude faster than LRTDP. In the SAP300 problems, the running times of all algorithms except TVI increase as search depth increases. LRTDP runs fast when $d$ is relatively small, but it slows down considerably and is unable to solve many problems when $d$ becomes larger. ILAO*'s convergence speed varies a bit when the distance is small. As $d$ increases, its running time also increases. BRTDP's performance (not included) is close to that of ILAO* when $d$ is small, but becomes slower and performs similar to LRTDP when $d$ is large. In this problem, heuristic search algorithms unanimously suffer significantly from the increase in the search depth, as their running times increase by at least two orders of magnitude from small to large $d$ values. On the other hand, FTVI slows down by only one order of magnitude, which makes it converge one order of magnitude faster than ILAO*, one to two orders of magnitude faster than BRTDP and TVI, and two orders of magnitude faster than LRTDP for large depths.

---

8. Note that these problems have well-defined initial states. Here we picked initial states arbitrarily from $\mathcal{S}$.

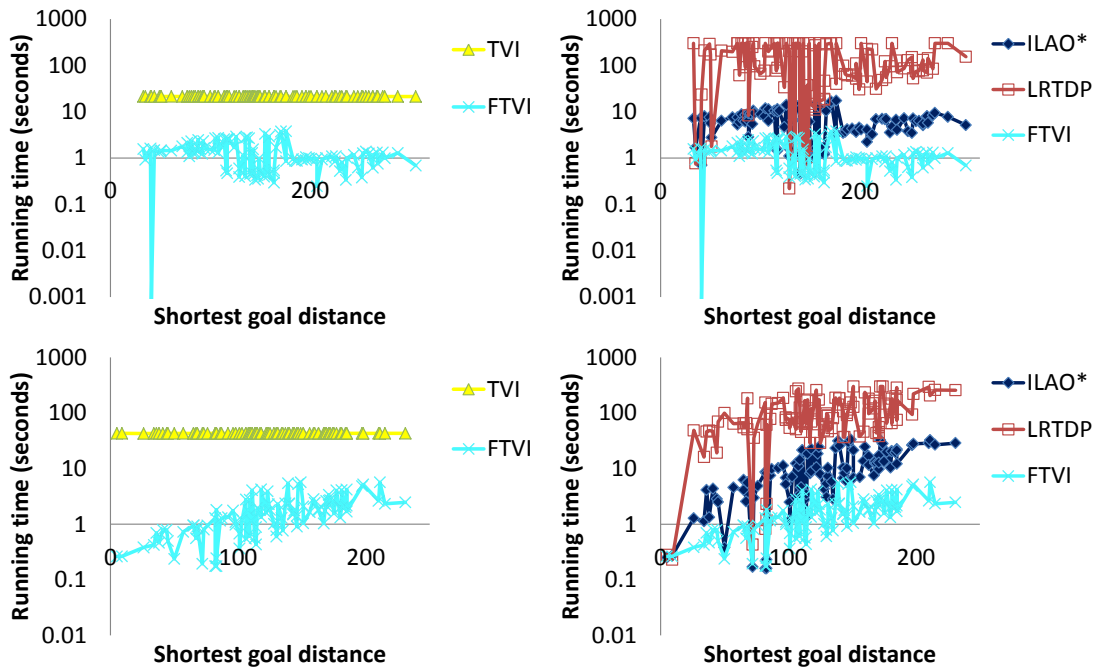

Figure 7: Running times of algorithms with different shortest distance to the goal on (top) mountain car $300 \times 300$ (MCar300), (bottom) single-arm pendulum $300 \times 300$ (SAP300) problems, (left) comparison of FTVI and TVI, and (right) comparison of FTVI and heuristic search algorithms. Heuristic search algorithms slow down massively (note the log scale) when the search depth is large.

*Heuristic Quality.* Finally we studied the effect of the heuristic informativeness on the algorithms. We conducted two sets of experiments, based on two sets of consistent heuristics. We found BRTDP slower than other algorithms in all problems and BaRTDP to be comparable (about 50% slower than LRTDP) only on the Wet100 problem, so did not include their running times. In the first experiment, we pre-computed the optimal value function of a problem using value iteration, and used a fraction of the optimal value as an initial heuristic. Given a fraction $f \in (0, 1]$, we calculated $h(s) = f \times V^*(s)$. Figure 8 plots the running times of different algorithms against $f$ for three problems. Note that $f = 1$ means the initial heuristic is already optimal, so a problem is trivial for all algorithms, but TVI has the overhead of building a topological structure. FTVI, however, is able to detect convergence in the search step and circumvent this overhead, so it is fast. LRTDP is slow in the Wet100 problem, so its running times in that problem are omitted from the figure. The figure shows that as $f$ increases (i.e. as the heuristic becomes more informative) the running times of all algorithms decrease almost linearly. This is true even for TVI, which is not a heuristic-guided algorithm, but takes less time, probably because the initial values affect the number of iterations required until convergence.

To thoroughly study the influence of the heuristics, we conducted a second set of experiments. In this experiment, we used a fractional $V_l$ value as our initial heuristic. Recall that $V_l$ is a lower bound of $V^*$ computed by the value of a deterministic problem. We calculated the initial heuristic by $h(s) = f \times V_l(s)$. All included algorithms show a similar smooth decrease in running time when $f$ increases. BRTDP, however, shows strong dependence on the heuristics in the Wet100 problem. Its running time decreases sharply from 96.91 seconds to 0.54 seconds and from 99.81 seconds to 6.21 seconds from when $f = 0.02$ to when $f = 1$ in the two experiments. Stable changes in the two experiments suggests the following for algorithms except BRTDP. (1) No algorithm is particularly vulnerable to a less informed heuristic function; (2) extremely informative heuristics (when $f$ is very close to 1) do not necessarily lead to extra fast convergence. This result is in-line with results for deterministic domains (Helmert & Röger, 2008).

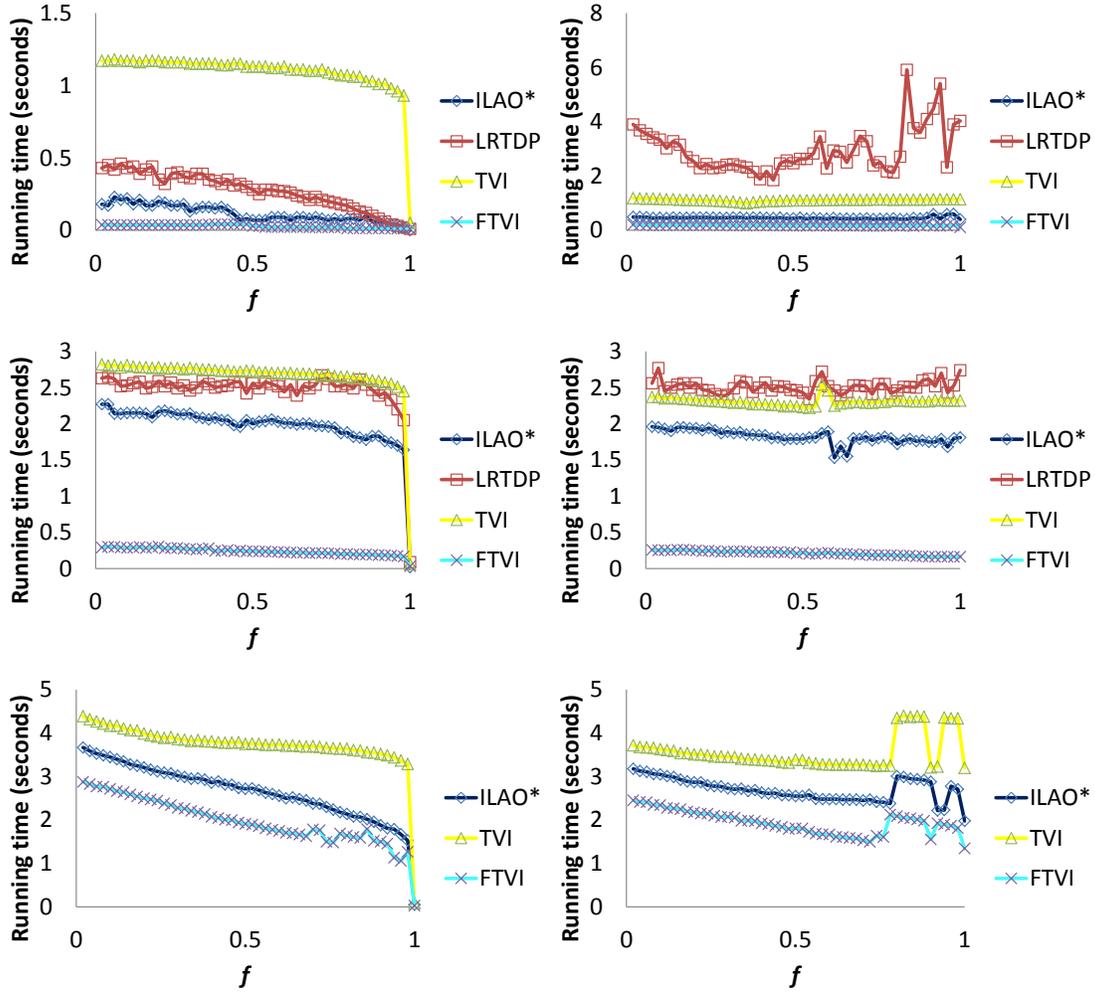

Figure 8: Running times of algorithms with different initial heuristic on (top) mountain car $100 \times 100$ (MCar100), (middle) single-arm pendulum $100 \times 100$ (SAP100), and (bottom) wet-floor $100 \times 100$ (WF100) problems. All algorithms are equally sensitive to the heuristic informativeness. (left) $f = \sum_{s \in \mathcal{S}} h(s) / \sum_{s \in \mathcal{S}} V^*(s)$ (right) $f = \sum_{s \in \mathcal{S}} h(s) / \sum_{s \in \mathcal{S}} V_l(s)$.

### 4.3.3 DISCUSSION

From the experiments, we learn that FTVI is vastly better in domains whose problems have a small number of goal states and a long search depth from the initial state to a goal (such as MCar, SAP and Drive). But the convergence control module of FTVI helps in successfully matching the performance of FTVI with the fastest heuristic search algorithm. In addition, FTVI displays limited advantage over heuristic search in the two intermediate cases where a problem has (1) many goal states but long search depth (Elevator), or (2) a short depth but fewer goal states (DAP). In conclusion, FTVI is our algorithm of choice whenever a problem has either a small number of goal states or a long search depth.

## 5. Related Work

Besides TVI several other researchers have proposed decomposing an MDP into sub-problems and combining their solutions for the final policy, e.g., the work of Hauskrecht et al. (1998) and Parr (1998). However, these approaches typically assume some additional structure in the problem, either known hierarchies, or known decomposition into weakly coupled sub-MDPs, etc., whereas FTVI assumes no additional structure.

BRTDP (McMahan et al., 2005), Bayesian RTDP (Sanner et al., 2009) and Focused RTDP (Smith & Simmons, 2006) (FRTDP) also keep an upper bound for the value function. However, all algorithms use the upper bound purely to judge how close a state is to convergence, by comparing the difference between the upper and lower bound values. For example, BRTDP tries to make searches focus more on states whose two bounds have larger differences, or intuitively, states whose values are less converged. Unlike FTVI, all three algorithms do not perform action elimination, nor do they use any connected component information to solve an MDP. The performance of BRTDP (and similarly Bayesian RTDP) is highly dependent on the quality of the heuristics. Furthermore, FRTDP only works for the discounted setting, thus is not immediately applicable for stochastic shortest path problems.

HDP is similar to TVI in the sense that it uses the Tarjan's algorithm (slightly different from the Kosaraju's algorithm) to find the strongly connected components of a greedy graph. It computes the SCCs multiple times and dynamically during the depth-first searches when HDP tries to label solved states. But it does not find the topological order of the SCCs nor decompose a problem and use the topological order to sequentially solve each SCC.

Prioritized sweeping (Moore & Atkeson, 1993) and its extensions, focussed dynamic programming (Ferguson & Stentz, 2004) and improved prioritized sweeping (McMahan & Gordon, 2005), order backups intelligently with the help of a priority queue. Each state in the queue is prioritized based on the potential improvement in value of a backup over that state. Dai and Hansen (2007) demonstrate that these algorithms have large overhead in maintaining a priority queue so they are outperformed by a simple backward search algorithm, which implicitly prioritizes backups without a priority queue. Moreover, prioritized sweeping and improved prioritized sweeping find the optimal value of the entire state space of an MDP, as they do not use the initial state information. Focussed dynamic programming, however, is able to make use of the initial state information, but it is not an optimal algorithm. All three algorithms are massively outperformed by an LAO* variant (Dai & Hansen, 2007).

When an MDP is too large to be solved optimally, another thread of work solves MDPs approximately. The typical way to do this is to use deterministic relaxations of the MDP and/or basis functions (Guestrin, Koller, Parr, & Venkataraman, 2003; Poupart, Boutilier, Patrascu, & Schuurmans, 2002; Patrascu, Poupart, Schuurmans, Boutilier, & Guestrin, 2002; Yoon, Fern, & Givan, 2007; Kolobov, Mausam, & Weld, 2009, 2010a, 2010b). The techniques of these algorithms are orthogonal to the ones by FTVI, and an interesting future direction is to approximate FTVI by applying basis functions.

When an MDP maintains a logical representation, another type of algorithm aggregates groups of states of an MDP by features, represents them as a factored MDP using algebraic and Boolean decision diagrams (ADDs and BDDs) and solves the factored MDP using ADD and BDD operations; SPUDD (Hoey, St-Aubin, Hu, & Boutilier, 1999), sLAO* (Feng & Hansen, 2002), sRTDP (Feng, Hansen, & Zilberstein, 2003) are examples. The factored representation can be exponentially simpler than a flat MDP, but the computation efficiency is problem-dependent. The idea of these algorithms are orthogonal to those of (F)TVI. Exploring ways of combining the ideas of (F)TVI with compact logical representation to achieve further performance improvements remains future work.

Action elimination was originally proposed by Bertsekas (2001). It has been proved to be helpful for RTDP in the factored MDP setting (Kuter & Hu, 2007), when the cost of an action depends on only a few state variables. Action elimination is also very useful in temporal planning (Mausam & Weld, 2008). It has been extended to *combo-elimination*, a rule to prune irrelevant action combinations in a setting when multiple actions can be executed at the same time.

The idea of finding the topological order of strongly connected components of an MDP has been extended to solving *partially-observable MDPs* (POMDPs). A POMDP problem is typically much harder than an MDP problem since the decision agent only has partial information of the current state (Littman et al., 1995). The topological order-based planner (POT) (Dibangoye, Shani, Chaib-draa, & Mouaddib, 2009) uses the topological order information of the underlying MDPs to help solve a POMDP problem faster. We believe the idea can be extended to help solve even harder problems, such as decentralized POMDP (Bernstein, Givan, Immerman, & Zilberstein, 2002), in the future.

## 6. Conclusions

This work makes several contributions. First, we present two new optimal algorithms to solve MDPs, topological value iteration (TVI) and focused topological value iteration (FTVI). TVI studies the graphical structure of an MDP by breaking it into strongly connected components and solves the MDP based on the topological order of the components. FTVI extends topological value iteration algorithm by focusing the construction of strongly connected components on transitions that likely belong to an optimal policy. FTVI does this by using a small amount of heuristic search to eliminate provably suboptimal actions. In contrast to TVI, which does not care about goal-state information, FTVI removes transitions which it determines to be irrelevant to an optimal policy for reaching the goal. In this sense, FTVI builds a much more informative topological structure than TVI.

Second, we show empirically that TVI outperforms VI and other state-of-the-art algorithms when an MDP contains many strongly connected components. We find that TVI is the most advantageous on problems with multiple equal-sized components.

Third, we show empirically that FTVI outperforms TVI and VI in a large number of domains, usually by an order of magnitude. This performance is due to the success of a more informed graphical structure, since the sizes of the connected components found by FTVI are vastly smaller than those constructed by TVI's.

Fourth, we find surprisingly that for many domains FTVI massively outperforms popular heuristic search algorithms in convergence speed, such as ILAO*, LRTDP, BRTDP and BaRTDP. After analyzing the performance of these algorithms over different problems, we find that a smaller number of goal states and long search depth to a goal are two key features of problems that are especially hard for heuristic search to handle. Our results show that FTVI outperforms heuristic search in such domains by an order of magnitude.

Finally, as a by-product we also compare ILAO*, LRTDP, BRTDP and BaRTDP (four popular, state-of-the-art heuristic search algorithms) and find that the strength of each algorithm is usually domain-specific. Generally, ILAO* is faster in convergence than other algorithms. BRTDP and BaRTDP are slow in some domains probably due to the fact that they are vulnerable to those problems' lack of informed upper bounds.

## Acknowledgments

This work was conducted when Peng Dai was a student at the University of Washington. This work was supported by Office of Naval Research grant N00014-06-1-0147, National Science Foundation IIS-1016465, ITR-0325063 and the WRF / TJ Cable Professorship. We thank Eric A. Hansen for sharing his code for ILAO*, and anonymous reviewers for excellent suggestions on improving the manuscript.

## References

Aberdeen, D., Thiébaux, S., & Zhang, L. (2004). Decision-Theoretic Military Operations Planning. In *Proc. of the 14th International Conference on Automated Planning and Scheduling (ICAPS-04)*, pp. 402–412.

Barto, A., Bradtke, S., & Singh, S. (1995). Learning to act using real-time dynamic programming. *Artificial Intelligence J.*, *72*, 81–138.

Bellman, R. (1957). *Dynamic Programming*. Princeton University Press, Princeton, NJ.

Bernstein, D. S., Givan, R., Immerman, N., & Zilberstein, S. (2002). The Complexity of Decentralized Control of Markov Decision Processes. *Mathematics of Opererations Research*, *27*(4), 819–840.

Bertsekas, D. P. (2000-2001). *Dynamic Programming and Optimal Control*, Vol. 2. Athena Scientific.

Bertsekas, D. P., & Tsitsiklis, J. N. (1996). *Neuro-Dynamic Programming*. Athena Scientific, Belmont, MA.

Bonet, B., & Geffner, H. (2003a). Faster Heuristic Search Algorithms for Planning with Uncertainty and Full Feedback. In *Proc. of 18th International Joint Conf. on Artificial Intelligence (IJCAI-03)*, pp. 1233–1238. Morgan Kaufmann.

Bonet, B., & Geffner, H. (2003b). Labeled RTDP: Improving the Convergence of Real-time Dynamic Programming. In *Proc. 13th International Conference on Automated Planning and Scheduling (ICAPS-03)*, pp. 12–21.

Bonet, B. (2006). Non-Deterministic Planning Track of the 2006 International Planning Competition.. http://www.ldc.usb.ve/˜bonet/ipc5/.

Bonet, B. (2007). On the Speed of Convergence of Value Iteration on Stochastic Shortest-Path Problems. *Mathematics of Operations Research, 32*(2), 365–373.

Bonet, B., & Geffner, H. (2006). Learning in Depth-First Search: A Unified Approach to Heuristic Search in Deterministic Non-deterministic Settings, and Its Applications to MDPs. In *Proc. of the 16th International Conference on Automated Planning and Scheduling (ICAPS-06)*, pp. 142–151.

Bresina, J. L., Dearden, R., Meuleau, N., Ramkrishnan, S., Smith, D. E., & Washington, R. (2002). Planning under Continuous Time and Resource Uncertainty: A Challenge for AI. In *Proc. of 18th Conf. in Uncertainty in AI (UAI-02)*, pp. 77–84.

Cormen, T. H., Leiserson, C. E., Rivest, R. L., & Stein, C. (2001). *Introduction to Algorithms, Second Edition*. The MIT Press.

Dai, P., & Goldsmith, J. (2007). Topological Value Iteration Algorithm for Markov Decision Processes. In *Proc. of IJCAI*, pp. 1860–1865.

Dai, P., & Hansen, E. A. (2007). Prioritizing Bellman Backups Without a Priority Queue. In *Proc. of the 17th International Conference on Automated Planning and Scheduling (ICAPS-07)*, pp. 113–119.

Dai, P., Mausam, & Weld, D. S. (2008). Partitioned External-Memory Value Iteration. In *AAAI*, pp. 898–904.

Dai, P., Mausam, & Weld, D. S. (2009a). Domain-Independent, Automatic Partitioning for Probabilistic Planning. In *IJCAI*, pp. 1677–1683.

Dai, P., Mausam, & Weld, D. S. (2009b). Focused Topological Value Iteration. In *Proc. of ICAPS*, pp. 82–89.

Dibangoye, J. S., Shani, G., Chaib-draa, B., & Mouaddib, A.-I. (2009). Topological Order Planner for POMDPs. In *Proc. of IJCAI*, pp. 1684–1689.

Feng, Z., & Hansen, E. A. (2002). Symbolic Heuristic Search for Factored Markov Decision Processes. In *Proc. of the 17th National Conference on Artificial Intelligence (AAAI-05)*.

Feng, Z., Hansen, E. A., & Zilberstein, S. (2003). Symbolic Generalization for On-line Planning. In *Proc. of the 19th Conference in Uncertainty in Artificial Intelligence (UAI-03)*, pp. 209–216.

Feng, Z., & Zilberstein, S. (2004). Region-Based Incremental Pruning for POMDPs. In *Proc. of UAI*, pp. 146–153.

Ferguson, D., & Stentz, A. (2004). Focussed Dynamic Programming: Extensive Comparative Results. Tech. rep. CMU-RI-TR-04-13, Carnegie Mellon University, Pittsburgh, PA.

Guestrin, C., Koller, D., Parr, R., & Venkataraman, S. (2003). Efficient Solution Algorithms for Factored MDPs. *J. of Artificial Intelligence Research, 19*, 399–468.

Hansen, E. A., & Zilberstein, S. (2001). LAO*: A heuristic search algorithm that finds solutions with loops. *Artificial Intelligence J.*, *129*, 35–62.

Hauskrecht, M., Meuleau, N., Kaelbling, L. P., Dean, T., & Boutilier, C. (1998). Hierarchical Solution of Markov Decision Processes using Macro-actions. In *Proc. of UAI*, pp. 220–229.

Helmert, M., & Röger, G. (2008). How Good is Almost Perfect?. In *Proc. of AAAI*, pp. 944–949.

Hoey, J., St-Aubin, R., Hu, A., & Boutilier, C. (1999). SPUDD: Stochastic Planning using Decision Diagrams. In *Proc. of the 15th Conference on Uncertainty in Artificial Intelligence (UAI-95)*, pp. 279–288.

Kolobov, A., Mausam, & Weld, D. S. (2009). ReTrASE: Intergating Paradigms for Approximate Probabilistic Planning. In *Proc. of IJCAI*, pp. 1746–1753.

Kolobov, A., Mausam, & Weld, D. S. (2010a). Classical Planning in MDP Heuristics: With a Little Help from Generalization. In *Proc. of ICAPS*, pp. 97–104.

Kolobov, A., Mausam, & Weld, D. S. (2010b). SixthSense: Fast and Reliable Recognition of Dead Ends in MDPs. In *Proc. of AAAI*.

Kuter, U., & Hu, J. (2007). Computing and Using Lower and Upper Bounds for Action Elimination in MDP Planning. In *SARA*, pp. 243–257.

Littman, M. L., Dean, T., & Kaelbling, L. P. (1995). On the Complexity of Solving Markov Decision Problems. In *Proc. of the 11th Annual Conference on Uncertainty in Artificial Intelligence (UAI-95)*, pp. 394–402 Montreal, Quebec, Canada.

Mausam, Benazera, E., Brafman, R. I., Meuleau, N., & Hansen, E. A. (2005). Planning with Continuous Resources in Stochastic Domains. In *Proc. of IJCAI*, pp. 1244–1251.

Mausam, & Weld, D. S. (2008). Planning with Durative Actions in Stochastic Domains. *J. of Artificial Intelligence Research (JAIR)*, *31*, 33–82.

McMahan, H. B., & Gordon, G. J. (2005). Fast Exact Planning in Markov Decision Processes. In *Proc. of the 15th International Conference on Automated Planning and Scheduling (ICAPS-05)*.

McMahan, H. B., Likhachev, M., & Gordon, G. J. (2005). Bounded real-time dynamic programming: RTDP with monotone upper bounds and performance guarantees. In *Proceedings of the 22nd international conference on Machine learning (ICML-05)*, pp. 569–576.

Meuleau, N., Benazera, E., Brafman, R. I., Hansen, E. A., & Mausam (2009). A Heuristic Search Approach to Planning with Continuous Resources in Stochastic Domains. *J. of Artificial Intelligence Research (JAIR)*, *34*, 27–59.

Moore, A., & Atkeson, C. (1993). Prioritized Sweeping: Reinforcement Learning with Less Data and Less Real Time. *Machine Learning*, *13*, 103–130.

Musliner, D. J., Carciofini, J., Goldman, R. P., E. H. Durfee, J. W., & Boddy, M. S. (2007). Flexibly Integrating Deliberation and Execution in Decision-Theoretic Agents. In *ICAPS Workshop on Planning and Plan-Execution for Real-World Systems*.

Nilson, N. J. (1980). *Principles of Artificial Intelligence*. Tioga Publishing Company, Palo Alto, Ca.

Parr, R. (1998). Flexible Decomposition Algorithms for Weakly Coupled Markov Decision Problems. In *Proc. of UAI*, pp. 422–430.

Patrascu, R., Poupart, P., Schuurmans, D., Boutilier, C., & Guestrin, C. (2002). Greedy Linear Value-Approximation for Factored Markov Decision Processes. In *Proc. of the 17th National Conference on Artificial Intelligence (AAAI-02)*, pp. 285–291.

Poupart, P., Boutilier, C., Patrascu, R., & Schuurmans, D. (2002). Piecewise Linear Value Function Approximation for Factored MDPs. In *Proc. of the 18th National Conference on Artificial Intelligence (AAAI-02)*, pp. 292–299.

Sanner, S., Goetschalckx, R., Driessens, K., & Shani, G. (2009). Bayesian Real-Time Dynamic Programming. In *Proc. of IJCAI*, pp. 1784–1789.

Smith, T., & Simmons, R. G. (2006). Focused Real-Time Dynamic Programming for MDPs: Squeezing More Out of a Heuristic. In *Proc. of the 21th National Conference on Artificial Intelligence (AAAI-06)*.

Wingate, D., & Seppi, K. D. (2005). Prioritization Methods for Accelerating MDP Solvers. *J. of Machine Learning Research*, *6*, 851–881.

Yoon, S., Fern, A., & Givan, R. (2007). FF-Replan: A Baseline for Probabilistic Planning. In *Proc. of the 17th International Conference on Automated Planning and Scheduling (ICAPS-07)*, pp. 352–359.